\documentclass{egpubl}
\usepackage{VMV2025}
\ArxivPaper
\usepackage[T1]{fontenc}
\usepackage{dfadobe}  
\usepackage{microtype}
\usepackage{enumitem}
\usepackage{collect}
\usepackage{xspace}
\usepackage{amsfonts}
\usepackage{amsmath}
\usepackage{stmaryrd}
\usepackage{mathtools}
\usepackage{soul}
\usepackage{booktabs}

\biberVersion
\BibtexOrBiblatex
\usepackage[backend=biber,bibstyle=EG,citestyle=alphabetic,backref=true]{biblatex} 
\electronicVersion
\PrintedOrElectronic

\ifpdf \usepackage[pdftex]{graphicx} \pdfcompresslevel=9
\else \usepackage[dvips]{graphicx} \fi

\usepackage{egweblnk}

\newcommand{\eg}{e.g.,\ }
\newcommand{\ie}{i.e.,\ }
\newcommand{\etal}{et~al.\ }

\newcommand{\refFig}[1]{Fig.~\ref{fig:#1}}
\newcommand{\refTab}[1]{Tab.~\ref{tab:#1}}
\newcommand{\refSec}[1]{Sec.~\ref{sec:#1}}
\newcommand{\refEq}[1]{Eq.~\ref{eq:#1}}

\newcommand{\method}[1]{\textsc{#1}}

\usepackage{xcolor}
\definecolor{changecolor}{rgb}{0.1, 0.8, 0.2} 
\newcommand{\change}[1]{{#1}}

\definecollection{mymaths}

\newcommand{\mymath}[2]{
    \newcommand{#1}{\TextOrMath{$#2$\xspace}{#2}}
    \begin{collect}{mymaths}{}{}{}{}
    #1
    \end{collect}
}

\mymath{\R}{\mathbb{R}}
\mymath{\fct}{f}
\mymath{\antiderivative}{F}
\mymath{\trainableparams}{{\boldsymbol{\theta}}}
\mymath{\field}{\antiderivative_\trainableparams}
\mymath{\mcantiderivative}{\hat{\antiderivative}}
\mymath{\coord}{x}
\mymath{\indimcount}{d}
\mymath{\outdimcount}{d_o}
\mymath{\domain}{\Omega}
\mymath{\intcount}{n}
\mymath{\samples}{\mathcal{S}}
\mymath{\mcsamplecount}{N}
\mymath{\uniformdist}{\mathcal{U}}
\mymath{\loss}{\mathcal{L}}
\mymath{\expectation}{\mathbb{E}}
\mymath{\diffop}{\mathcal{D}}
\mymath{\compop}{\mathcal{B}}
\mymath{\reductionfactor}{\alpha}
\mymath{\reducedantiderivative}{\Phi}
\mymath{\examplefct}{g}
\mymath{\eps}{\epsilon}
\mymath{\smoothkernel}{\kappa}
\mymath{\convcoord}{\tau}
\mymath{\refantiderivative}{\antiderivative_\text{ref}}
\mymath{\randominput}{\xi}
\mymath{\mcfct}{\hat{f}}

\title{Learning Neural Antiderivatives}

\author{Submission 1007}

\author[Rubab et al.]
{
F. Rubab$^{1, 2}$\orcid{0009-0001-6979-7746},		
N. E. Nsampi$^1$\orcid{0000-0002-3769-9850}, 
M. Balint$^1$\orcid{0000-0001-6689-4770},
F. Mujkanovic$^1$\orcid{0009-0009-9122-4408},
H.-P. Seidel$^1$\orcid{0000-0002-1343-8613},	
T. Ritschel$^3$\orcid{0009-0006-4660-7790}	and
T. Leimkühler$^1$\orcid{0009-0006-7784-7957}
\\
$^1$ Max-Planck-Institut für Informatik\qquad
$^2$ Michigan State University\qquad
$^3$ University College London
}

\begin{document}


\maketitle

\begin{abstract}
Neural fields offer continuous, learnable representations that extend beyond traditional discrete formats in visual computing.
We study \change{the problem of learning neural representations of repeated antiderivatives directly from a function, a} continuous analogue of summed-area tables.
Although widely used in discrete domains, such cumulative schemes rely on grids, which prevents their applicability in continuous neural contexts.
We introduce and analyze a range of neural methods for repeated integration, including both adaptations of prior work and novel designs. 
Our evaluation spans multiple input dimensionalities and integration orders, assessing both reconstruction quality and performance in downstream tasks such as filtering and rendering.
These results \change{enable} integrating classical cumulative operators into modern neural systems and offer insights into learning tasks involving differential and integral operators.

\ccsdesc[500]{Computing methodologies~Machine learning algorithms}
\ccsdesc[300]{Computing methodologies~Image manipulation}
\ccsdesc[300]{Computing methodologies~Rendering}

\printccsdesc   
\end{abstract}  


\section{Introduction and Related Work}

Neural representations are playing an increasingly prominent role in modeling core primitives in visual computing. 
Moving beyond traditional, specialized formats -- such as pixel grids for images or triangle meshes for surface geometry -- neural fields~\cite{xie2022neural} offer a unified and continuous representation by mapping spatial coordinates to function values~\cite{functa22}.
These representations are not hand-crafted, but learned from data via gradient-based optimization.
Especially important are cases where the available data originate in a different domain than the one being reconstructed, with the two connected by a differentiable forward model, enabling the solution of inverse problems.
Popular examples of this paradigm include learning a scene’s appearance from image observations~\cite{mildenhall2020nerf}, where the forward model is volume rendering, and learning a signed distance function (SDF) from depth maps~\cite{liu2020dist}, where the forward model is sphere tracing.

In this work, we consider a specific, yet highly fundamental, inverse problem to be addressed using neural fields: learning repeated antiderivatives. 
Specifically, given, and given only, training samples from a function with one or more input variables, our goal is to obtain a representation that encodes the result of integration of that function -- possibly multiple times and along multiple dimensions. \change{Unlike a single integration which only yields box filtering, repeated integration enables weighted aggregation through higher-order splines, as we show in applications.}

In the context of discrete representations, this problem has been the subject of extensive research in computer graphics, computer vision, and adjacent fields. 
Summed-area tables (SAT)~\cite{crow1984summed} -- also known as prefix sums or integral images -- are regular grids that store the (repeated) cumulative sum of an underlying tabulated function, and constitute a fundamental data structure.
\change{
While SATs work in the pixel basis, pre-integration of other basis functions is possible, such as for (moving) least squares \cite{levin1998approximation} (Example 2), for splines \cite{ahlberg1967theory} or subdivision surfaces \cite{CATMULL1978350}.
We would not be aware of a general way to do this for neural representations.
}
Over the decades, SATs have supported a wide range of applications, including 
filtering~\cite{heckbert1986filtering,bradley2007adaptive}, 
pattern matching~\cite{viola2001rapid,veksler2003fast,ipol.2014.57,bay2006surf,porikli2005integral,wu2024fast}, 
rendering~\cite{hensley2005fast,hensley2007fast,lauritzen2007summed,li2021multi}, 
media streaming~\cite{li2021log}, neural architecture design~\cite{babiloni2023adaptive,babiloni2023factorized,yalavarthi2025efficient}, 
and multiscale modeling~\cite{zhu2024realistic,li2023continuouslodlfn,atanasov2021multiscale}.
The efficient computation of cumulative sums has emerged as a productive research area in its own right~\cite{blelloch1989scans,sengupta2007scan,nehab2011gpu,kasagi2014parallel,chaurasia2015compiling,Urschler_2013_ICCV_Workshops}.
However, the inherent dependence on regular grids renders all these methods unsuitable for continuous neural representations.

Motivated by recent advances in the signal-processing capabilities of neural fields~\cite{yang2021geometry,xu2022signal,Nsampi2023NeuralFC,mujkanovic2024ngssf}, as well as their success in solving inverse problems involving differential equations~\cite{raissi2019physics,chakraborty2022computing}, we investigate the neural, continuous analogue of summed-area tables: learning repeated integrals from sampled data.
After reviewing the necessary mathematical preliminaries~(\refSec{preliminaries}), we define the problem setting in \refSec{setting}.
Although analytic solutions exist for obtaining single-integral functions using a highly restricted class of shallow neural architectures~\cite{subr2021q,lloyd2020using}, scalable and broadly applicable approaches require dedicated learning strategies.
To this end, we provide a hierarchical taxonomy of methods~(\refSec{strategies}, \refFig{taxonomy}) -- some adapted from existing techniques originally developed in different contexts, and others representing novel contributions to the design space.
We  evaluate each class of methods across modalities with varying input dimensionalities and integration orders, employing a range of computational tools -- including assessments of antiderivative quality on synthetic and real data, as well as performance evaluation in downstream tasks such as filtering and rendering~(\refSec{evaluation}).
We conclude with concrete recommendations (\refSec{conclusion}).

In summary, our contributions are:
\begin{itemize}
    \item A taxonomy of methods for learning repeated antiderivatives from function samples, including novel method designs.
    \item An evaluation of these methods across signal modalities, input dimensionalities, and integration orders.
    \item A demonstration of the practical impact of method choice in downstream tasks such as filtering and rendering.
\end{itemize}

All source code and trained models are available at \url{https://neural-antiderivatives.mpi-inf.mpg.de/}.


\section{Preliminaries}
\label{sec:preliminaries}

In this section, we review mathematical concepts relevant to our task and its solutions.
Specifically, we recall techniques for reducing repeated integrals (\refSec{prelim-integral-reduction}) and provide a discussion of gradient bias in stochastic estimators (\refSec{prelim-gradient-bias}).


\subsection{Integral Reduction}
\label{sec:prelim-integral-reduction}

Repeated integrals of a function
$
\fct(\coord): \R \rightarrow \R
$
of the form
\begin{equation}
    \int^{(\intcount)}
    \!\! \fct (\coord) \,\,
    \mathrm{d}\coord^{(\intcount)}
    \coloneq
    \underbrace{
        \int \!\! \ldots \! \int
    }_{\intcount \text{ times}}
    \fct (\coord) \,\,
    \underbrace{
        \vphantom\int
        \mathrm{d}\coord \ldots \mathrm{d}\coord
    }_{\intcount \text{ times}}
\end{equation}
can be reduced to equivalent formulations involving only single integrals.
We recall two techniques in this context:
The classical formula of Cauchy~\cite{cauchy1823infinitesimal}
\begin{equation}
\label{eq:cauchy}
    \int^{(\intcount)}
    \!\! \fct (\coord) \,\,
    \mathrm{d}\coord^{(\intcount)}
    =
    \frac{1}{(\intcount-1)!}
    \int_{-\infty}^{\coord}
    (\coord - \coord')^{\intcount-1}
    \fct (\coord') \,\,
    \mathrm{d}\coord'
\end{equation}
and a recent formulation by Haddad~\cite{haddad2021repeated}
\begin{equation}
\label{eq:haddad}
     \int^{(\intcount)} \!\! 
     \fct (\coord) \,\,
    \mathrm{d}\coord^{(\intcount)}
    =
    \sum_{l=1}^\intcount
    \reductionfactor_{\intcount, l} \,\,
    \coord^{l-1}
    \int
    \coord^l \fct (\coord) \,\,
    \mathrm{d}\coord,
\end{equation}
which uses scalar coefficients 
$
\reductionfactor_{\intcount, l}
=
\frac
{(-1)^{\intcount - l}}
{(l-1)!(\intcount - l)!}.
$
There are two fundamental differences between these reduction formulas.
First, \refEq{cauchy} collapses the repeated integral into a single integral, whereas \refEq{haddad} rewrites it as a sum of \intcount separate single integrals.
Second, the individual summands in \refEq{haddad} are genuine antiderivatives, whereas the integrand in \refEq{cauchy} depends on the upper integration limit -- \ie it is an explicit function of the evaluation coordinate \coord.
In other words, the reduction in Equation~\refEq{cauchy} yields a spatially varying single integral, whereas the reduction in Equation~\refEq{haddad} results in a sum of stationary antiderivatives.


\subsection{Gradient Bias of Stochastic Estimators}
\label{sec:prelim-gradient-bias}

Consider the stochastic computation 
$\fct_\trainableparams(\randominput)$, 
where \randominput denotes a random input to a deterministic model $\fct_\trainableparams$, which is parameterized by trainable parameters \trainableparams.
Our objective is to learn \trainableparams via gradient descent, using a loss function \loss, \ie we are interested in computing
\begin{equation}
\label{eq:true-gradient}
    \nabla_\trainableparams
    \loss
    \left(
        \expectation_\randominput
        \left[
            \fct_\trainableparams(\randominput)
        \right]
    \right).
\end{equation}
In practice, computing the expectation
$\expectation_\randominput \left[ \fct_\trainableparams(\randominput) \right]$
is infeasible and therefore approximated using a Monte Carlo estimator $\mcfct_\trainableparams$. 
Unfortunately, the resulting expected gradient
\begin{equation}
\label{eq:actual-gradient}
    \expectation_\randominput
    \left[
        \nabla_\trainableparams
        \loss
        \left(    
                \mcfct_\trainableparams
        \right)
    \right]
\end{equation}
is generally biased with respect to the true gradient defined in \refEq{true-gradient}, even when $\mcfct_\trainableparams$ is itself an unbiased estimator, for two reasons.
First, the interchange of the loss gradient and the expectation in the transition from \refEq{true-gradient} to \refEq{actual-gradient} is valid only when the loss gradient is a linear function of its argument.
This condition is naturally satisfied only by the squared error (L2) loss~\cite{nicolet2023recursive}; any other loss function leads to biased gradient estimates.
The second source of bias becomes apparent upon applying the chain rule to expand \refEq{actual-gradient}:
\begin{equation}
\label{eq:chain-rule}
     \expectation_\randominput
    \left[
        \nabla_\trainableparams
        \loss
        \left(    
                \mcfct_\trainableparams
        \right)
    \right]
    =
    \expectation_\randominput
    \left[
        \frac
        {
            \partial \loss \left( \mcfct_\trainableparams \right)
        }
        {
            \partial \mcfct_\trainableparams
        }
        \cdot
        \nabla_\trainableparams \mcfct_\trainableparams
    \right].
\end{equation}
Both the outer and inner derivatives involve the same random variable $\mcfct_\trainableparams$.
Since the expectation of a product equals the product of the expectations only when the random variables are independent, this dependence introduces a bias~\cite{azinovic2019inverse}, whose magnitude scales with the variance of $\mcfct_\trainableparams$~\cite{deng2022reconstructing}.

In practice, gradient bias in stochastic models can be mitigated using two countermeasures:
First, by employing the L2 loss; and second, by explicitly computing the product in \refEq{chain-rule} using estimators $\mcfct_\trainableparams$ with different random inputs \randominput for each factor.


\section{Setting}
\label{sec:setting}

We consider the setting where a given signal
$
\fct(\coord_1, \dots, \coord_\indimcount)
: 
\R^{\indimcount} \rightarrow \R^{\outdimcount}
$
can be sampled, but is otherwise unknown.
\change{Sampling is possible at any arbitrary position, hence we do not solve a (scattered) data interpolation problem \cite{levin1998approximation}, neither do we extrapolate, but purely seek to represent a signal and its antiderivative.}
We assume that \(\fct\) \change{defined on the unit hyper-cube} 
$\domain = [0, 1]^\indimcount$ \change{and leave analysis on other domains, \eg 3D surface manifolds to future work}.
The samples are denoted by
\begin{equation}
\label{eq:samples}
    \samples = \left\{ \left( \coord_{1, i}, \dots, \coord_{\indimcount, i}, \fct_i \right) \right\}_i,    
\end{equation}
where 
$
\fct_i \coloneq \fct(\coord_{1, i}, \dots, \coord_{\indimcount, i})
$. 
Our goal is to obtain the continuous, repeated antiderivative
$
\antiderivative^{(\intcount)}(\coord_1, \dots, \coord_\indimcount)
:
\R^{\indimcount} \rightarrow \R^{\outdimcount},
$
\ie the result of applying \intcount-fold integration to \fct along each of the \indimcount input dimensions, resulting in a total of 
$\indimcount \times \intcount$ 
integration steps:
\begin{equation}
\label{eq:integral}
    \antiderivative^{(\intcount)}(\coord_1, \dots, \coord_\indimcount)
    \coloneq
    \int^{(\intcount)} \!\!\!\! \ldots \! \int^{(\intcount)}
    \!\! \fct (\coord_1, \dots, \coord_\indimcount) \,\,
    \mathrm{d}\coord_1^{(\intcount)} \ldots \mathrm{d}\coord_{\indimcount}^{(\intcount)}.
\end{equation}
To support notational clarity and a consistent evaluation structure, we assume that all \indimcount input dimensions are integrated the same number of times, \intcount.
This cases arises most frequently in practical applications.
However, all formulations and analysis results naturally extend to the more general case where the number of integrations may vary across dimensions.
In our setting, we further ignore the (higher-order) constants of integration -- a single integral is defined up to a constant, a double integral up to a constant and a linear term, and so on -- as these terms cancel out exactly in practical applications of $\antiderivative^{(\intcount)}$.

While numerical integration would, in principle, allow the computation of $\antiderivative^{(\intcount)}$ at any specific coordinate $(\coord_1, \dots, \coord_\indimcount)$ in isolation~\cite{caflisch1998monte}, our objective is to learn the full continuous antiderivative as a unified model
$
\field^{(\intcount)}(\coord_1, \dots, \coord_\indimcount)
\approx 
\antiderivative^{(\intcount)}(\coord_1, \dots, \coord_\indimcount),
$
parameterized by \trainableparams,
which enables efficient evaluation at arbitrary coordinates.
To this end, we employ a neural field~\cite{xie2022neural} to represent $\field^{(\intcount)}$, due to its flexibility in modeling diverse modalities and dimensionalities.

In the remainder of this work, we explore different methods for the non-trivial task of learning $\field^{(\intcount)}$ from \samples.
In this context, we largely disregard the influence of specific neural architectures, as their impact proved to be minor compared to that of the learning strategy.
After a reasonably comprehensive pilot architecture search, we adopted a Multilayer Perceptron (MLP) with Swish activation functions~\cite{ramachandran2017searching} and axis-aligned positional encoding~\cite{mildenhall2020nerf,tancik2020fourier} for all our experiments -- an architecture that has also been identified as effective in a related integration setting~\cite{lindell2021autoint}.
The network has four hidden layers with 256 neurons each and is trained using the Adam~\cite{kingma2014adam} optimizer. \change{We use the default Adam parameters with a learning rate of $10^{-3}$.} 
To isolate the effect of different supervision structures, we refrain from varying the loss function except where strictly necessary (\refSec{prelim-gradient-bias}).
A pilot study indicated that the Huber loss yields the most consistent performance across integration orders, input dimensionalities, and signal modalities. \change{Training is run for $100{,}000$ iterations for methods based on automatic differentiation and direct integral supervision, and for $200{,}000$ iterations when numerical differentiation is employed. Batch sizes vary depending on the integration order, input dimensionality, and method, due to time and memory considerations.}


\section{Methods}
\label{sec:strategies}

\begin{figure}
    \includegraphics[width=\linewidth]{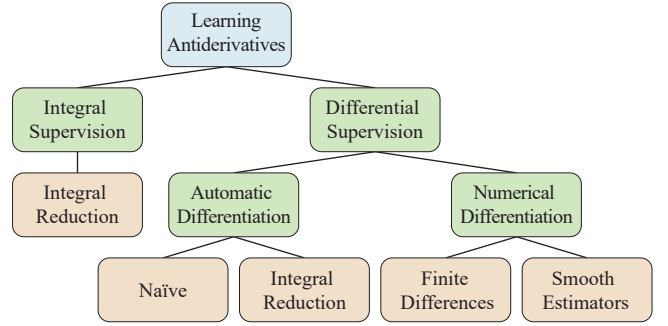}
    \caption{
        Overview of our method taxonomy.
        The task of learning antiderivatives (blue) can be addressed using method classes (green), each comprising specific algorithmic realizations (orange).
        \change{For definitions of each algorithmic realization, see Sec.~\ref{sec:Methods}.}
    }
    \label{fig:taxonomy}
\end{figure}

\begin{figure*}
    \includegraphics[width=\linewidth]{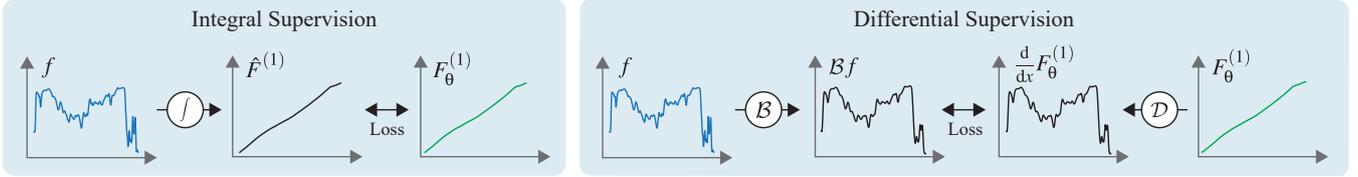}
    \caption{
        Overview of the two classes of approaches to learning repeated antiderivatives $\antiderivative^{(\intcount)}$ (green) from function samples, illustrated for a single integration of a one-dimensional function \fct (blue).
        Left: Integral supervision involves numerically estimating antiderivative values $\mcantiderivative^{(\intcount)}$ across the domain and using this estimate to guide training.
        Right: Differential supervision begins by applying the differential operator \diffop to the model.
        The resulting signal is then compared to the original function samples, optionally modified by a compensation operator \compop to account for approximation errors introduced by \diffop.
        Different choices of \diffop lead to methods with varying trade-offs between accuracy and computational efficiency.
    }
    \label{fig:overview}
\end{figure*}

Here, we present a taxonomy of training methods designed to learn a the (repeated) antiderivative $\field^{(\intcount)}$, given the samples \samples. 
See \refFig{taxonomy} for the taxonomy tree and \refFig{overview} for graphical illustrations.

\subsection{Integral Supervision}
\label{sec:strategy-integral}

In this method, the samples \samples are used to numerically estimate the target antiderivative $\antiderivative^{(\intcount)}$ at specific coordinates, and the model $\field^{(\intcount)}$ is trained to match this estimate (\refFig{overview}, left).
Since our objective is to compute a \emph{repeated} integral, this naturally gives rise to a nested Monte Carlo estimator~\cite{rainforth2018nesting}.
However, we observe that such an estimator fails to converge to reasonable antiderivative estimates under practical compute constraints, due to the nesting induced by repeated integration.
We therefore turn to integral reduction. 
Although both formulas in \refSec{prelim-integral-reduction} are compatible with Monte Carlo sampling, Cauchy’s reduction (\refEq{cauchy}) involves only a single integration. 
Consequently, we adopt this formulation for our task:
\begin{equation}
\begin{split}
    &\antiderivative^{(\intcount)}(\coord_1, \dots, \coord_\indimcount)
    \approx
    \mcantiderivative^{(\intcount)}(\coord_1, \dots, \coord_\indimcount) \\
    \coloneq
    &\frac
    {\prod_{j=1}^{\indimcount} \coord_j}
    {\mcsamplecount (\intcount-1)!}
    \sum_{k=1}^{\mcsamplecount}
    \prod_{j=1}^{\indimcount}
    \left(
        \coord_j - \coord_{j,k}
    \right)^{\intcount-1}
    \fct(\coord_{1,k}, \dots, \coord_{\indimcount,k}),
\end{split}
\end{equation}
where \mcsamplecount denotes the number of Monte Carlo samples, and 
$\coord_{j,k} \sim \uniformdist \left[0, \coord_j \right]$,
\ie we draw samples uniformly from the \indimcount-dimensional, axis-aligned hyperrectangle defined by the origin and the current coordinate $(\coord_1, \dots, \coord_\indimcount)$.
In practice, convergence is improved by using \indimcount-dimensional Sobol sequences~\cite{sobol1967distribution} to generate the sample locations.
The estimate $\mcantiderivative^{(\intcount)}$ can then be used to directly train the model $\field^{(\intcount)}$ via the loss
\begin{equation}
    \loss
    =
    \expectation_{(\coord_1, \dots, \coord_\indimcount) \in \domain}
    \left[
    \left\|
         \field^{(\intcount)}(\coord_1, \dots, \coord_\indimcount)
         -
         \mcantiderivative^{(\intcount)}(\coord_1, \dots, \coord_\indimcount)
    \right\|
    \right].
\end{equation}


\subsection{Differential Supervision}
\label{sec:strategy-differential}

At their core, differential supervision strategies are grounded in the fundamental theorem of calculus~\cite{stewart2012calculus}, which, when applied to our setting, takes the form
\begin{equation}
\label{eq:fundamentaltheorem}
    \frac
    {
        \partial^{\indimcount \intcount}
    }
    {
        \partial \coord_1^\intcount \dots \partial \coord_\indimcount^\intcount
    }
    \antiderivative^{(\intcount)}
    =
    \fct,
\end{equation}
\ie integration and differentiation act as inverse operations.
This motivates a class of training strategies whose loss takes the general form
\begin{equation}
\label{eq:differential-general}
    \loss
    =
    \mathbb{E}_{(\coord_1, \dots, \coord_\indimcount) \in \domain}
    \left[
    \left\|
        \diffop\,
        \field^{(\intcount)}(\coord_1, \dots, \coord_\indimcount)
        -
        \compop\,
        \fct(\coord_1, \dots, \coord_\indimcount)
    \right\|
    \right],
\end{equation}
where the operator
\begin{equation}
\label{eq:diff-definition}
    \diffop
    \approx
    \frac
    {
        \partial^{\indimcount \intcount}
    }
    {
        \partial \coord_1^\intcount \dots \partial \coord_\indimcount^\intcount
    }
\end{equation}
approximates the repeated differentiation described in \refEq{fundamentaltheorem}, and \compop is a method-specific operator introduced to compensate for errors introduced by the approximation in \diffop (\refFig{overview}, right).
An appealing property of this class of methods is that it bypasses the need for \emph{pre-integration} using samples \samples, as required in \refSec{strategy-integral}, and instead makes use of \samples in a more direct supervision setting involving the \emph{differentiated model}.
Since all operations involved are local, this approach typically offers substantial efficiency gains compared to the global estimators required for numerical integration.

\begin{figure}
    \includegraphics[width=\linewidth]{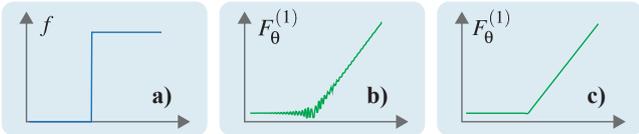}
    \caption{
        An example illustrating the effect of the compensation operator \compop in \refEq{differential-general}, when using a differential operator \diffop that introduces blurring.  
        (\emph{a}) The function \fct, accessible through samples, is a 1D step function.  
        (\emph{b}) Training the first antiderivative $\field^{(1)}$ without compensating for the blur introduced by \diffop results in ringing artifacts that contaminate the solution.  
        (\emph{c}) Applying an appropriate compensation operator \compop eliminates the artifacts and yields a faithful solution.
    }
    \label{fig:compensation-didactical}
\end{figure}

To understand why the compensation operator \compop is necessary, consider the common case where \diffop does not compute the exact repeated derivative, but instead produces a blurred approximation.
\refFig{compensation-didactical} illustrates the effect of training with the loss from \refEq{differential-general}, when $\compop$ is set to the identity operator.
We observe that $\field^{(\intcount)}$ fails to converge to an accurate approximation of the true antiderivative, instead exhibiting pronounced ringing artifacts (\refFig{compensation-didactical}b).
The artifacts can be eliminated by applying the blur introduced by \diffop to the solution as well, effectively canceling the ringing.
To avoid such a costly post-processing step, we instead incorporate the blur into \fct via the compensation operator \compop in \refEq{differential-general}, resulting in a more faithful solution (\refFig{compensation-didactical}c).
In practice, this smoothing is estimated using a Monte Carlo convolution~\cite{hermosilla2018monte}.
Note that this approach can yield slightly blurred antiderivatives, as both the model output and the supervision signal are smoothed -- providing little to no incentive for the model to capture high-frequency details.  
Nevertheless, a mildly blurred solution, as shown in \refFig{compensation-didactical}c, is typically preferable to the more severe artifacts observed in \refFig{compensation-didactical}b.

Furthermore, note that losses of the form \refEq{differential-general} admit infinitely many minimizers, corresponding to models $\field^{(\intcount)}$ that differ by (higher-order) constants of integration, which vanish under differentiation -- an invariance with no practical impact.

In the remainder of this section, we present specific realizations of \refEq{differential-general}, corresponding to particular choices of \diffop and \compop.


\subsubsection{Automatic Differentiation}
\label{sec:strategy-automatic-diff}

The most straightforward way to apply the repeated differential operator \diffop to a neural model $\field^{(\intcount)}$ is through automatic differentiation~\cite{linnainmaa1970representation,rumelhart1985learning}, which is natively supported by all major machine learning frameworks~\cite{paszke2017automatic,bradbury2018jax,abadi2016tensorflow}.
Automatic differentiation of the model output with respect to its input constructs or augments a computational graph that evaluates the derivative directly, using either forward-mode or reverse-mode variants.
This derivative graph shares the trainable parameters \trainableparams with the original model and can be used for training~\cite{teichert2019machine,lindell2021autoint}.
Notably, this approach yields an exact equality in \refEq{diff-definition}, up to numerical inaccuracies.
As a result, the compensation operator \compop can be safely set to the identity.
Following Lindell~\etal~\cite{lindell2021autoint}, we use a normalized positional encoding to improve training stability.

While implementations of higher-order automatic differentiation vary, the memory and computational complexity of the resulting derivative graphs typically scale superlinearly with the order of differentiation.
Consequently, performing $\indimcount \times \intcount$ differentiations sequentially and backpropagating gradients through the resulting graphs becomes increasingly demanding as the order increases.


\paragraph{Integral Reduction}
\label{sec:integral-reduction}

Integral reduction (\refSec{prelim-integral-reduction}) offers a way to curb the explosive growth of the computational graph caused by repeated applications of automatic differentiation. 
Accordingly, we adopt Haddad’s formulation (\refEq{haddad}) --  Cauchy’s reduction (\refEq{cauchy}) does not yield antiderivatives, which are required for differential supervision.
Letting 
$\left\{ \reducedantiderivative_l \right\}$ 
denote the first-order antiderivatives in \refEq{haddad}, we write
\begin{equation}
\label{eq:haddad-combination}
    \antiderivative^{(\intcount)}(\coord)
    =
    \sum_{l=1}^\intcount
    \reductionfactor_{\intcount, l} \,\,
    \coord^{l-1}
    \reducedantiderivative_l(\coord).
\end{equation}
The key advantage of this formulation is that, instead of applying \emph{higher}-order automatic differentiation to a \emph{single} model $\field^{(\intcount)}$, we apply \emph{first}-order automatic differentiation to \intcount models 
$\left\{ \reducedantiderivative_{\trainableparams, l} \right\}$ using the loss
\begin{equation}
    \loss
    =
    \expectation_{\coord \in \domain}
    \left[
    \sum_{l=1}^\intcount
    \left\|
        \frac{\mathrm{d}}{\mathrm{d} \coord}
         \reducedantiderivative_{\trainableparams, l}(\coord)
         -
         \coord^l \fct(\coord)
    \right\|
    \right].
\end{equation}
After training, the models 
$\left\{ \reducedantiderivative_{\trainableparams, l} \right\}$ 
are combined according to \refEq{haddad-combination}.
Extension to multiple dimensions is achieved by applying \refEq{haddad-combination} recursively, resulting in $\intcount^\indimcount$ models, each requiring only \indimcount automatic differentiations during training.

Since the individual models $\left\{ \reducedantiderivative_{\trainableparams, l} \right\}$ are highly correlated -- their derivatives differ only by powers of the input coordinates -- we found that training a single model with $\intcount^\indimcount$ outputs performs comparably to training separate models.
Evidently, this approach requires minor architectural modifications to the neural field, specifically in the form of a wider output layer.


\subsubsection{Numerical Differentiation}

Unlike automatic differentiation, which leverages the structure of a model’s computational graph, numerical differentiation estimates derivatives by evaluating the model at perturbed input coordinates~\cite{burden2015numerical}.
Crucially, it requires only forward passes through the model, and is therefore often significantly more efficient -- particularly for larger $\indimcount$ and $\intcount$.
However, replacing computations of instantaneous rates of change with finite coordinate perturbations to approximate \diffop naturally introduces a smoothing effect, which can be addressed through an appropriate choice of \compop.
Note that integral reduction (\refSec{integral-reduction}) could, in principle, also be used in combination with numerical differentiation; however, unlike in the case of automatic differentiation, it provides no efficiency advantage.

In the following, we present the two commonly used classes of techniques for learning with numerical differentiation.
Our exposition focuses on the 1D case, as the extension to higher-dimensional operators can be readily achieved using outer products of 1D coordinate perturbation offsets.


\paragraph{Finite Differences}
\label{sec:strategy-fd}

A widely used approach for estimating the derivative is the scheme
\begin{equation}
\label{eq:finite-difference}
    \frac{\mathrm{d}}{\mathrm{d} \coord}
    \examplefct (\coord)
    \approx
    \frac
    {
        \examplefct(\coord + \eps) 
        -
        \examplefct(\coord - \eps)
    } 
    {2 \eps},
\end{equation}
where $\eps \in \R_{+}$ denotes a small, fixed coordinate perturbation.
This central difference operation can be interpreted as the convolution of \examplefct with a pair of Dirac delta functions.
It is the the natural and minimal approach for numerically estimating derivatives in continuous representations that support sampling at arbitrary coordinates~\cite{li2023neuralangelo,Nsampi2023NeuralFC} -- unlike in discrete representations, where it is common to choose among different additional schemes (\eg forward and backward differences) to balance trade-offs such as bias and numerical stability.

\begin{figure}
    \includegraphics[width=\linewidth]{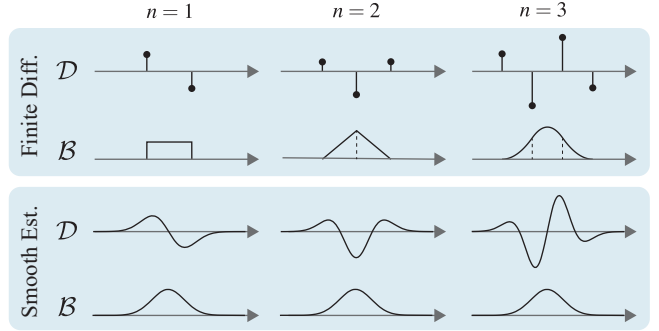}
    \caption{
        Convolution kernels for numerical differentiation operators \diffop and the corresponding compensation operators \compop, shown across different orders of differentiation \intcount.  
        In simple finite differencing, the kernel is a set of Dirac delta functions (first row).
        This introduces smoothing via piecewise polynomial kernels (second row; dashed lines indicate the boundaries of the segments).
        Smooth estimators use derivatives of the Gaussian distribution (third row), which inherently introduce smoothing through the undifferentiated Gaussian kernel (last row).
    }
    \label{fig:fd-sampling}
\end{figure}

Higher-order derivatives can be estimated by recursively applying \refEq{finite-difference}, resulting in the sampling schemes illustrated in the first row of \refFig{fd-sampling}.  
Since sampling locations coincide across recursive steps, the number of required samples grows linearly rather than exponentially:  
computing a derivative of order \intcount requires $\intcount + 1$ samples.
The compensation operation \compop corresponding to this scheme is a convolution with a piecewise polynomial blur kernel~\cite{heckbert1986filtering, Nsampi2023NeuralFC}, as illustrated in the second row of \refFig{fd-sampling}.

\paragraph{Smooth Estimators}
\label{sec:strategy-smooth}

The scheme in \refEq{finite-difference}, along with its higher-order extensions, suffers from a key limitation: 
the coordinate perturbation \eps must be predetermined and remains fixed.
This is of only minor concern in discrete representations, where \eps typically corresponds to the signal's sample spacing, thereby ensuring that little to no information is lost.
However, continuous neural representations tend to allocate spatial resolution adaptively.
Therefore, a fixed perturbation magnitude may result in an aliased estimate of the derivative~\cite{shannon1949communication}.

A solution to this problem is to replace the convolution with Dirac delta functions with a convolution using the (higher-order) derivative of a smooth, non-negative kernel \smoothkernel~\cite{fischer2023plateau,fischer2024zerograds}:
\begin{equation}
\label{eq:smooth-derivative}
    \frac{\mathrm{d}^\intcount}{\mathrm{d} \coord^\intcount}
    \examplefct (\coord)
    \approx
    \int
    \frac{\mathrm{d}^\intcount}{\mathrm{d} \convcoord^\intcount}
    \smoothkernel
    (\convcoord) 
    \,
    \examplefct(\coord - \convcoord)
    \,
    \mathrm{d} \convcoord.
\end{equation}
A natural choice for \smoothkernel is the Gaussian distribution, which admits analytic expressions for its (higher-order) derivatives (third row in \refFig{fd-sampling}).
In practice, \refEq{smooth-derivative} is computed using the Monte Carlo estimate
\begin{equation}
\label{eq:mc-convolution}
    \frac{\mathrm{d}^\intcount}{\mathrm{d} \coord^\intcount}
    \examplefct (\coord)
    \approx
    \frac{1}{\mcsamplecount_\smoothkernel}
    \sum_{\convcoord_i}
    \frac{\mathrm{d}^\intcount}{\mathrm{d} \convcoord^\intcount}
    \smoothkernel
    (\convcoord_i) 
    \,
    \examplefct(\coord - \convcoord_i),
\end{equation}
using $\mcsamplecount_\smoothkernel$ random samples $\convcoord_i$.
Kernel derivatives 
$\frac{\mathrm{d}^\intcount}{\mathrm{d}\convcoord^\intcount} \smoothkernel$
with odd \intcount yield odd functions (first and third column in \refFig{fd-sampling}), for which antithetic sampling~\cite{hammersley1956general} is highly effective in reducing the variance of the estimate.
Furthermore, low-discrepancy sampling using Sobol sequences~\cite{sobol1967distribution} has a significant positive effect on convergence.
Regardless of \intcount, the compensation operator \compop associated with this scheme corresponds to a convolution with \smoothkernel itself (last row in \refFig{fd-sampling}).

A variation of this scheme involves using stochastic, Gaussian-weighted sampling to fit a local \intcount-th-order polynomial to $g$, from which derivative values can be read off directly~\cite{Chetan_2025_CVPR}.
\refEq{mc-convolution} is a stochastic estimator of the form given in \refEq{actual-gradient}.
In practice, we observe that a na\"{i}ve implementation results in substantial gradient bias, to the point where the results become practically unusable.
Therefore, we apply the de-biasing techniques discussed in \refSec{prelim-gradient-bias}.


\section{Evaluation}
\label{sec:evaluation}

In this section, we present a systematic evaluation of all methods introduced above.
We begin by describing our evaluation protocol (\refSec{protocol}), followed by a presentation of the experimental results (\refSec{outcome}).


\subsection{Protocol}
\label{sec:protocol}

\subsubsection{Data}

We select a representative set of signals \fct, spanning input dimensionalities \indimcount from one to three, organized into two general categories.
First, we include isolated function classes composed of synthetic signals. 
For each \indimcount, we use 
a random mixture of Gaussians to represent smooth behavior,  
a random mixture of hyper-rectangles to model discontinuities,  
and the Ackley function~\cite{ackley} to capture oscillatory characteristics.

\begin{figure}
    \includegraphics[width=\linewidth]{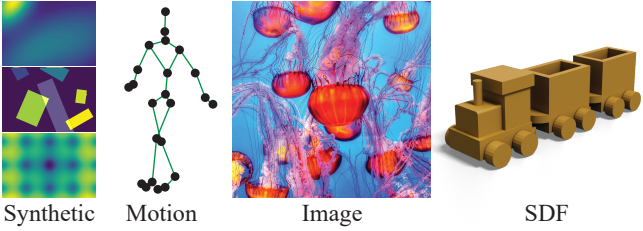}
    \caption{
        Example inputs from the set of test signals used in our evaluation, spanning multiple input dimensionalities.
    }
    \label{fig:data-samples}
\end{figure}

Second, for each \indimcount, we include a  set of representative real-world signals relevant to visual computing.
As a 1D representative, we use 20 motion-capture sequences, each consisting of 23 3D joint positions ($\outdimcount = 69$) capturing full-body human motion over 3.3 minutes.
As a 2D representative, we use 20 natural RGB images ($\outdimcount = 3$) with a resolution of $1640 \times 1640$, randomly sampled from the DIV2K dataset~\cite{Agustsson_2017_CVPR_Workshops}.
As a 3D representative, we use 10 signed distance functions (SDFs) ($\outdimcount = 1$) representing surface geometry, randomly drawn from multiple standard datasets~\cite{curless1996volumetric,Bogo_CVPR_FAUST_2014,chang2015shapenet,Thingi10K,Koch_2019_CVPR}.
All signals were padded to ensure proper boundary handling.
For the rendering task (\refSec{tasks}), we use three HDR environment maps at a resolution of $1024 \times 512$ sourced from an online repository~\cite{polyhaven}, each used to illuminate three meshes.
\refFig{data-samples} shows data samples.

\subsubsection{Tasks and Metrics}
\label{sec:tasks}

We learn the \intcount-fold antiderivative of the \indimcount-dimensional data, with \intcount ranging from one to three for $\indimcount=1$, and from one to two for $\indimcount > 1$.
We assess performance using two complementary approaches.
The first approach assesses antiderivative accuracy via reconstruction.
To this end, we apply (higher-order) automatic differentiation to the predicted antiderivatives and compare the result to the original data, since (higher-order) constants of integration may differ between solutions.
The second approach evaluates performance across different integration orders on two downstream tasks: filtering and rendering.
For the filtering task, we use the antiderivative representations of all real-world signals to perform convolution with a piecewise polynomial approximation of a \indimcount-dimensional Gaussian filter at varying bandwidths~\cite{heckbert1986filtering,Nsampi2023NeuralFC}.
The results are compared against converged Monte Carlo-based convolutions applied to the original data.
For the rendering task, we use antiderivative representations of environment maps to compute glossy reflections of distant illumination~\cite{hensley2005fast,hensley2007fast}.
Renderings from three different viewpoints are compared against corresponding results produced using the original environment maps.

We quantify errors using the mean squared error (MSE) across all signals.
For image outputs, we additionally report DSSIM~\cite{wang2004image} and LPIPS~\cite{zhang2018perceptual} scores, which incorporate structural similarity.
For all metrics, lower values indicate higher output quality.

\subsubsection{Methods}
\label{sec:Methods}
Our comparative evaluation includes all methods described in \refSec{strategies}:
\begin{itemize}
\item \method{Integral} refers to the integral supervision technique based on integral reduction (\refSec{strategy-integral}).
\item \method{AD-Na\"{i}ve} denotes differential supervision via automatic differentiation (\refSec{strategy-automatic-diff}).
\item \method{AD-Reduc} denotes automatic differentiation combined with integral reduction (\refSec{integral-reduction}).
\item \method{Num-FD} and \method{Num-FD-\compop} refer to numerical differentiation using finite differences, without and with the compensation operator \compop, respectively (\refSec{strategy-fd}).
\item \method{Num-Sm} and \method{Num-Sm-\compop} refer to numerical differentiation using smooth estimators, without and with the compensation operator \compop, respectively (\refSec{strategy-smooth}).
\end{itemize}


\subsection{Outcome}
\label{sec:outcome}

\begin{figure*}
    \includegraphics[width=\linewidth]{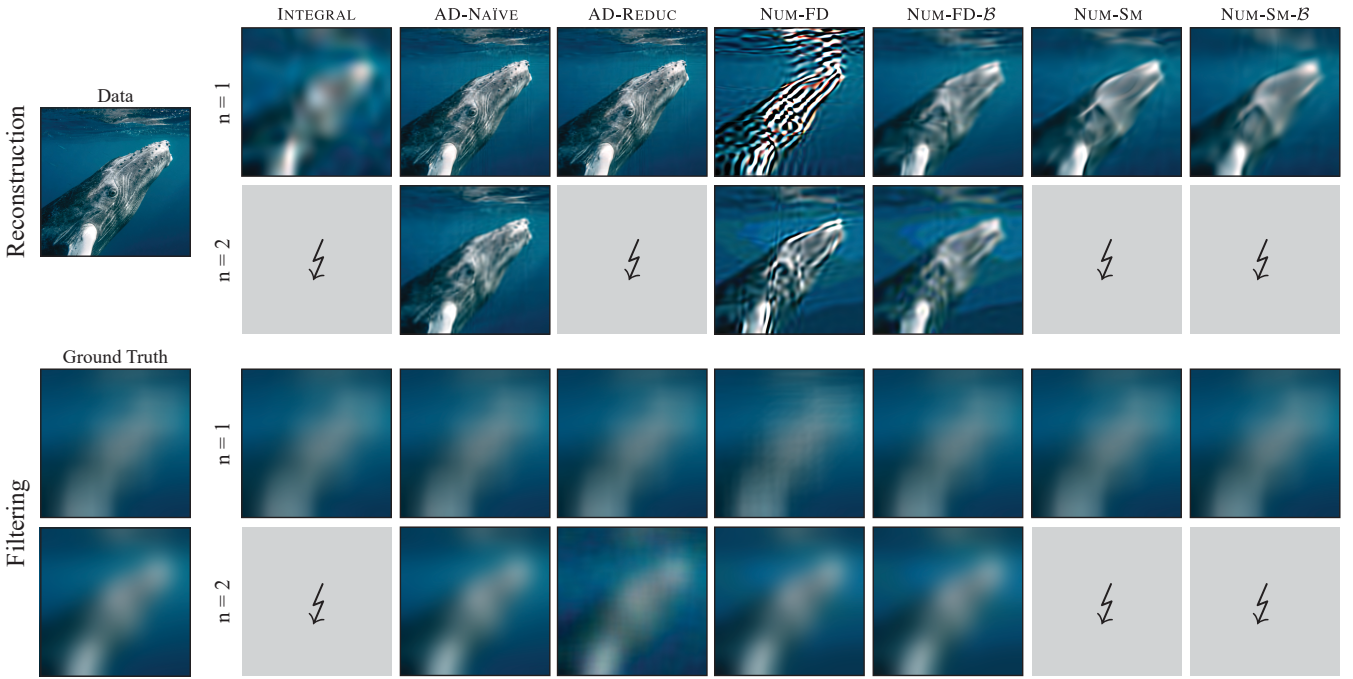}
    \caption{
        Qualitative results on images.
        The upper section illustrates reconstruction quality: the outputs of each method (columns) are shown after applying $2\intcount$-fold automatic differentiation to the learned antiderivatives.
        The lower section presents filtering results using a piecewise constant ($\intcount = 1$) and a piecewise linear ($\intcount = 2$) approximation of a Gaussian kernel.
        The symbol $\lightning$ indicates failure (see main text).
    }
    \label{fig:image-results}
\end{figure*}

\begin{figure}
    \includegraphics[width=\linewidth]{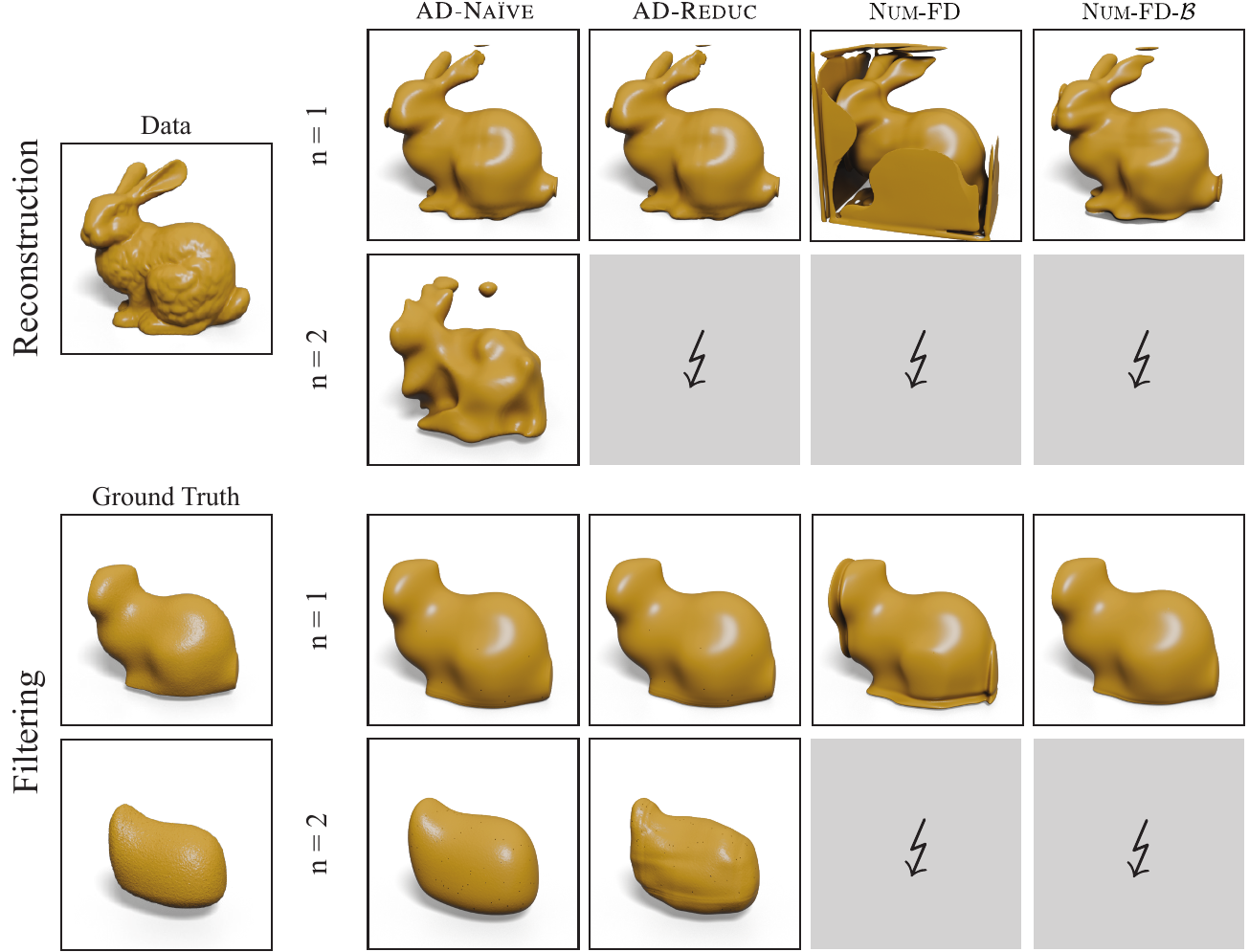}
    \vspace{-0.5cm}
    \caption{
        Qualitative results on SDFs.
        The upper section illustrates reconstruction quality: the outputs of each method (columns) are shown after applying $3\intcount$-fold automatic differentiation to the learned antiderivatives.
        The lower section presents filtering results using a piecewise constant ($\intcount = 1$) and a piecewise linear ($\intcount = 2$) approximation of a Gaussian kernel.
        The symbol $\lightning$ indicates failure (see main text); methods not shown failed for this modality.
    }
    \label{fig:sdf-results}
\end{figure}

\begin{figure*}
    \includegraphics[width=\linewidth]{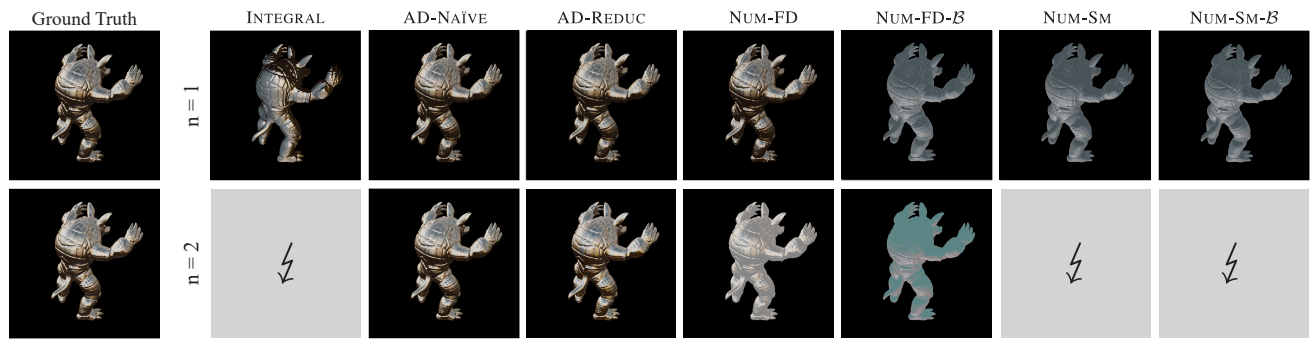}
    \caption{
        Qualitative rendering results.
        We present glossy image-based lighting results using a piecewise constant ($\intcount = 1$) and a piecewise linear ($\intcount = 2$) approximation of a Phong lobe.
        The symbol $\lightning$ indicates failure (see main text).
    }
    \label{fig:rendering-results}
\end{figure*}

The results of our quantitative evaluation are reported in
\refTab{results-synthetic} for the synthetic functions,
in \refTab{results-real} for the real-world signals,
in \refTab{results-filtering} for the filtering task, and
in \refTab{results-rendering} for the rendering task.
Qualitative results are shown in \refFig{image-results} for images, in \refFig{sdf-results} for SDFs, and in \refFig{rendering-results} for the rendering task.
We additionally report training times for each real-world signal modality in Table~\refTab{training-time}.

Consistent with expectations, our first observation is that increasing the product $\indimcount \times \intcount$ -- and thus the number of integrations -- makes the task more difficult for all methods.
Indeed, some methods fail to succeed across all tasks.
We do not report numbers for methods that completely fail on a specific sub-task -- either by failing to converge during training or by producing numerical results more than 100$\times$ worse in MSE than those of the best-performing method -- and mark these results with a $\lightning$ instead.

We begin by examining the antiderivative reconstruction results.
We observe that \method{Integral} performs well only on the simplest task, where $\indimcount = \intcount = 1$, whereas in all other cases, training fails to converge reliably or produces high errors due to excessive variance in the training signal.
\method{AD-Na\"{i}ve} consistently achieves the best performance across tasks and integration orders.
In fact, it is the only method that does not fail on any of the tasks.
However, this level of quality comes at the cost of significantly longer training times, often exceeding those of all competing methods by orders of magnitude.
In contrast, \method{AD-Reduc} performs reasonably well on a small selection of functions but completely fails on nearly all others.
We observe that training the individual sub-integrals
$\left\{ \reducedantiderivative_{\trainableparams, l} \right\}$
converges reliably and efficiently. 
However, their combination introduces interference patterns caused by slight misalignments between the sub-integrals.
Numerical differentiation in \method{Num-FD} and \method{Num-FD-\compop} yields reasonable results with moderate training times. \change{However, \method{Num-FD} can exhibit ringing artifacts due to the oscillatory frequency response of the finite-difference operator. The blur compensation operator \compop suppresses these artifacts, resulting in \method{Num-FD-\compop} significantly outperforming \method{Num-FD}.}
The smooth operators in \method{Num-Sm} and \method{Num-Sm-\compop} frequently failed to converge meaningfully, except in low-dimensional settings.

In the filtering task, \method{Integral} yields decent results for $\intcount = 1$, but becomes unreliable for higher-order integration, with quality degrading significantly as the filter kernel size increases.
\method{AD-Na\"{i}ve} again performs best overall, although its error increases with filter kernel size.
\method{AD-Reduc} performs well for larger kernels and higher integration orders, occasionally even outperforming \method{AD-Na\"{i}ve} in the higher-order regime.
\method{Num-FD-\compop} performs significantly better than \method{Num-FD}, reaching quality close to \method{AD-Na\"{i}ve}.
Again, \method{Num-Sm} and \method{Num-Sm-\compop} fail to converge reliably.

The results of the rendering task reveal a similar overall pattern.
\method{AD-Na\"{i}ve} consistently outperforms all other methods across glossiness levels and integration orders.
\method{AD-Reduc} performs decently but deteriorates rapidly as the glossiness level increases.
\method{Num-FD} offers a middle ground in terms of quality, while \method{Num-FD-\compop} performs surprisingly poorly in the HDR setting compared to its performance in the equivalent LDR image filtering scenario discussed previously.
We observe that the compensation operator \compop struggles to handle high-intensity highlights in the environment map when operating under a limited sample budget.

\change{
Overall, a good number of combinations is marked as to not have converged, and one could ask why this is.
Often, some minima are also poor, and our criteria to call something `not converged' or just report a high error, is arbitrary.
Unfortunately, no better differentiation might be available.
We experimented with several strategies to improve convergence, such as different network architectures, loss functions with additional constraints, and gradient debiasing, but none improved it significantly. Our experimental design cannot differentiate if this is because the gradients are `wrong`, if this is because they are noisy, if this particular optimizer cannot find the minimum of that cost function in general or a combination of all the above.
This is not a property of learning of antiderivatives, but a property of learning itself.
}

\begin{table*}[]
    \centering
    \small
    \renewcommand{\tabcolsep}{0.11cm}
    \caption{Quantitative evaluation of antiderivative quality on \textbf{synthetic functions}. We report the mean squared error (MSE) across input dimensionalities \indimcount and integration orders \intcount for different methods (shown in rows).
    The symbol $\lightning$ indicates failure (see main text).}
    \label{tab:results-synthetic}
    \begin{tabular}{lrrrrrrrrrrrrrrrrrrrrr}
        \toprule
        Class & \multicolumn{7}{c}{Gaussian Mixture} & \multicolumn{7}{c}{Rectangle Mixture} & \multicolumn{7}{c}{Ackley} \\
        \cmidrule(lr){2-8}
        \cmidrule(lr){9-15}
        \cmidrule(lr){16-22}
        \indimcount & \multicolumn{3}{c}{1} & \multicolumn{2}{c}{2} & \multicolumn{2}{c}{3} & \multicolumn{3}{c}{1} & \multicolumn{2}{c}{2} & \multicolumn{2}{c}{3} & \multicolumn{3}{c}{1} & \multicolumn{2}{c}{2} & \multicolumn{2}{c}{3} \\
        \cmidrule(lr){2-4}
        \cmidrule(lr){5-6}
        \cmidrule(lr){7-8}
        \cmidrule(lr){9-11}
        \cmidrule(lr){12-13}
        \cmidrule(lr){14-15}
        \cmidrule(lr){16-18}
        \cmidrule(lr){19-20}
        \cmidrule(lr){21-22}
        \intcount & \multicolumn{1}{c}{1} & \multicolumn{1}{c}{2} & \multicolumn{1}{c}{3} & \multicolumn{1}{c}{1} & \multicolumn{1}{c}{2} & \multicolumn{1}{c}{1} & \multicolumn{1}{c}{2} & \multicolumn{1}{c}{1} & \multicolumn{1}{c}{2} & \multicolumn{1}{c}{3} & \multicolumn{1}{c}{1} & \multicolumn{1}{c}{2} & \multicolumn{1}{c}{1} & \multicolumn{1}{c}{2} & \multicolumn{1}{c}{1} & \multicolumn{1}{c}{2} & \multicolumn{1}{c}{3} & \multicolumn{1}{c}{1} & \multicolumn{1}{c}{2} & \multicolumn{1}{c}{1} & \multicolumn{1}{c}{2}\\
        \addlinespace[0.25em]
        Scale &
        \tiny{$10^{-9}$} & \tiny{$10^{-8}$} & \tiny{$10^{-7}$} & \tiny{$10^{-7}$} & \tiny{$10^{-6}$} & \tiny{$10^{-7}$} & \tiny{$10^{-4}$} & \tiny{$10^{-4}$} & \tiny{$10^{-4}$} & \tiny{$10^{-3}$} & \tiny{$10^{-4}$} & \tiny{$10^{-3}$} & \tiny{$10^{-3}$} & \tiny{$10^{-2}$} & \tiny{$10^{-5}$} & \tiny{$10^{-6}$} & \tiny{$10^{-6}$} & \tiny{$10^{-7}$} & \tiny{$10^{-5}$} & \tiny{$10^{-6}$} & \tiny{$10^{-3}$} \\
        \midrule
        \method{Integral} &
        $\lightning$ & $\lightning$ & $\lightning$ & $\lightning$ & $\lightning$ & $\lightning$ & $\lightning$ & 89.7 & $\lightning$ & $\lightning$ & 108.4 & $\lightning$ & 70.4 & $\lightning$ & 2.2 & $\lightning$ & $\lightning$ & $\lightning$ & $\lightning$ & $\lightning$ & $\lightning$ \\
        \method{AD-Na\"{i}ve} &
        7.0 & 7.4 & 6.4 & 0.0 & 3.8 & 1.6 & 6.5 & 3.2 & 8.7 & 1.4 & 4.1 & 1.9 & 5.4 & 3.6 & 1.9 & 2.0 & 8.0 & 9.1 & 12.6 & 7.6 & 4.5 \\
        \method{AD-Reduc} &
        7.0 & $\lightning$ & $\lightning$ & 0.0 & $\lightning$ & 1.6 & $\lightning$ & 3.2 & $\lightning$ & $\lightning$ & 4.1 & $\lightning$ & 5.4 & $\lightning$ & 1.9 & 74.3 & $\lightning$ & 9.1 & $\lightning$ & 7.6 & $\lightning$ \\
        \method{Num-FD} &
        $\lightning$ & $\lightning$ & $\lightning$ & $\lightning$ & 1.2 & $\lightning$ & $\lightning$ & $\lightning$ & $\lightning$ & 20.1 & $\lightning$ & $\lightning$ & 279.5 & $\lightning$ & $\lightning$ & $\lightning$ & $\lightning$ & $\lightning$ & 22.4 & 546.5 & $\lightning$ \\
        \method{Num-FD-\compop} &
        275.0 & 74.1 & $\lightning$ & $\lightning$ & 1.1 & 69.2 & $\lightning$ & 19.4 & 80.1 & 14.6 & 19.3 & 3.0 & 9.8 & $\lightning$ & 1.2 & 8.2 & $\lightning$ & 47.2 & 2.2 & 18.2 & $\lightning$ \\
        \method{Num-Sm} &
        $\lightning$ & $\lightning$ & $\lightning$ & $\lightning$ & $\lightning$ & $\lightning$ & $\lightning$ & $\lightning$ & $\lightning$ & $\lightning$ & 80.4 & $\lightning$ & $\lightning$ & $\lightning$ & 25.5 & $\lightning$ & $\lightning$ & $\lightning$ & $\lightning$ & $\lightning$ & $\lightning$ \\
        \method{Num-Sm-\compop} &
        $\lightning$ & $\lightning$ & $\lightning$ & $\lightning$ & $\lightning$ & $\lightning$ & $\lightning$ & 46.1 & 442.3 & $\lightning$ & 7.3 & $\lightning$ & $\lightning$ & $\lightning$ & 16.9 & $\lightning$ & $\lightning$ & $\lightning$ & $\lightning$ & $\lightning$ & $\lightning$ \\
        \bottomrule 
    \end{tabular}
\end{table*}

\begin{table*}[]
    \small
    \caption{Quantitative evaluation of antiderivative quality across integration orders \intcount on \textbf{real-world functions}, using motion capture sequences, images, and signed distance functions (SDFs) as representative signals in visual computing. The symbol $\lightning$ indicates failure (see main text).}
    \label{tab:results-real}
    \begin{tabular}{lrrrrrrrrrrr}
        \toprule
        Modal. \tiny{($\indimcount \rightarrow \outdimcount$)} & \multicolumn{3}{c}{Motion \tiny{($1 \rightarrow 69$)}} & \multicolumn{6}{c}{Image \tiny{($2 \rightarrow 3$)}} & \multicolumn{2}{c}{SDF \tiny{($3 \rightarrow 1$)}} \\
        \cmidrule(lr){2-4}
        \cmidrule(lr){5-10}
        \cmidrule(lr){11-12}
        \intcount & 
        \multicolumn{1}{c}{1} & \multicolumn{1}{c}{2} & \multicolumn{1}{c}{3} & 
        \multicolumn{3}{c}{1} & \multicolumn{3}{c}{2} & 
        \multicolumn{1}{c}{1} & \multicolumn{1}{c}{2} \\
        \cmidrule(lr){2-2}
        \cmidrule(lr){3-3}
        \cmidrule(lr){4-4}
        \cmidrule(lr){5-7}
        \cmidrule(lr){8-10}
        \cmidrule(lr){11-11}
        \cmidrule(lr){12-12}
        Metric & \multicolumn{1}{c}{\tiny{MSE}} & \multicolumn{1}{c}{\tiny{MSE}} & \multicolumn{1}{c}{\tiny{MSE}} & 
        \multicolumn{1}{c}{\tiny{MSE}} & \multicolumn{1}{c}{\tiny{DSSIM}} & \multicolumn{1}{c}{\tiny{LPIPS}} & \multicolumn{1}{c}{\tiny{MSE}} & \multicolumn{1}{c}{\tiny{DSSIM}} & \multicolumn{1}{c}{\tiny{LPIPS}} &
        \multicolumn{1}{c}{\tiny{MSE}} & \multicolumn{1}{c}{\tiny{MSE}} \\
        \addlinespace[0.25em]
        Scale &
        \tiny{$10^{-4}$} & \tiny{$10^{-4}$} & \tiny{$10^{-4}$} & \tiny{$10^{-3}$} & \tiny{$10^{-1}$} & \tiny{$10^{-1}$} & \tiny{$10^{-3}$} & \tiny{$10^{-1}$} & \tiny{$10^{-1}$} & \tiny{$10^{-4}$} & \tiny{$10^{-3}$} \\
        \midrule
        \method{Integral} &
        11.5 & 152.5 & $\lightning$ & 25.6 & 3.2 & 8.7 & $\lightning$ & $\lightning$ & $\lightning$ & $\lightning$ & $\lightning$ \\
        \method{AD-Na\"{i}ve} &
        1.1 & 4.1 & 9.7 & 5.9 & 1.6 & 4.5 & 9.3 & 2.3 & 6.6 & 1.4 & 9.0 \\
        \method{AD-Reduc} &
        1.1 & $\lightning$ & $\lightning$ & 5.9 & 1.6 & 4.5 & $\lightning$ & $\lightning$ & $\lightning$ & 1.4 & $\lightning$ \\
        \method{Num-FD} &
        $\lightning$ & 277.8 & 53.4 & $\lightning$ & $\lightning$ & $\lightning$ & 847.9 & 3.3 & 7.9 & 29.4 & $\lightning$ \\
        \method{Num-FD-\compop} &
        5.1 & 6.1 & 49.5 & 13.3 & 2.7 & 7.6 & 15.3 & 3.0 & 8.3 & 5.5 & $\lightning$ \\
        \method{Num-Sm} &
        24.9 & $\lightning$ & $\lightning$ & 22.4 & 3.0 & 7.7 & $\lightning$ & $\lightning$ & $\lightning$ & $\lightning$ & $\lightning$ \\
        \method{Num-Sm-\compop} &
        1.2 & 218.7 & $\lightning$ & 3.0 & 1.0 & 3.3 & $\lightning$ & $\lightning$ & $\lightning$ & $\lightning$ & $\lightning$ \\
        \bottomrule 
    \end{tabular}
\end{table*}

\begin{table*}[]
    \centering
    \small
    \renewcommand{\tabcolsep}{0.07cm}
    \caption{Quantitative results for the use of learned antiderivatives of real-world signals in the task of \textbf{filtering}. We report results for two Gaussian kernel sizes -- small (S) and large (L) -- corresponding to standard deviations $\sigma = 0.1$ and $\sigma = 0.3$, respectively. The symbol $\lightning$ indicates failure (see main text).}
    \label{tab:results-filtering}
    \begin{tabular}{lrrrrrrrrrrrrrrrrrrrrrr}
        \toprule
        Modal. \tiny{($\indimcount \rightarrow \outdimcount$)} & \multicolumn{6}{c}{Motion \tiny{($1 \rightarrow 69$)}} & \multicolumn{12}{c}{Image \tiny{($2 \rightarrow 3$)}} & \multicolumn{4}{c}{SDF \tiny{($3 \rightarrow 1$)}} \\
        \cmidrule(lr){2-7}
        \cmidrule(lr){8-19}
        \cmidrule(lr){20-23}
        \intcount &
        \multicolumn{2}{c}{1}  & \multicolumn{2}{c}{2} & \multicolumn{2}{c}{3} &
        \multicolumn{6}{c}{1} & \multicolumn{6}{c}{2} &
        \multicolumn{2}{c}{1} & \multicolumn{2}{c}{2} \\
        \cmidrule(lr){2-3}
        \cmidrule(lr){4-5}
        \cmidrule(lr){6-7}
        \cmidrule(lr){8-13}
        \cmidrule(lr){14-19}
        \cmidrule(lr){20-21}
        \cmidrule(lr){22-23}
        Kernel Size & \multicolumn{1}{c}{S} & \multicolumn{1}{c}{L} & \multicolumn{1}{c}{S} & \multicolumn{1}{c}{L} & \multicolumn{1}{c}{S} & \multicolumn{1}{c}{L} &
        \multicolumn{3}{c}{S} & \multicolumn{3}{c}{L} & \multicolumn{3}{c}{S} & \multicolumn{3}{c}{L} & \multicolumn{1}{c}{S} & \multicolumn{1}{c}{L} & \multicolumn{1}{c}{S} & \multicolumn{1}{c}{L} \\
        \cmidrule(lr){2-2}
        \cmidrule(lr){3-3}
        \cmidrule(lr){4-4}
        \cmidrule(lr){5-5}
        \cmidrule(lr){6-6}
        \cmidrule(lr){7-7}
        \cmidrule(lr){8-10}
        \cmidrule(lr){11-13}
        \cmidrule(lr){14-16}
        \cmidrule(lr){17-19}
        \cmidrule(lr){20-20}
        \cmidrule(lr){21-21}
        \cmidrule(lr){22-22}
        \cmidrule(lr){23-23}
        Metric & \multicolumn{1}{c}{\tiny{MSE}} & \multicolumn{1}{c}{\tiny{MSE}} & \multicolumn{1}{c}{\tiny{MSE}} & \multicolumn{1}{c}{\tiny{MSE}} & \multicolumn{1}{c}{\tiny{MSE}} & \multicolumn{1}{c}{\tiny{MSE}} & 
        \multicolumn{1}{c}{\tiny{MSE}} & \multicolumn{1}{c}{\tiny{DSSIM}} & \multicolumn{1}{c}{\tiny{LPIPS}} & \multicolumn{1}{c}{\tiny{MSE}} & \multicolumn{1}{c}{\tiny{DSSIM}} & \multicolumn{1}{c}{\tiny{LPIPS}} & \multicolumn{1}{c}{\tiny{MSE}} & \multicolumn{1}{c}{\tiny{DSSIM}} & \multicolumn{1}{c}{\tiny{LPIPS}} & \multicolumn{1}{c}{\tiny{MSE}} & \multicolumn{1}{c}{\tiny{DSSIM}} & \multicolumn{1}{c}{\tiny{LPIPS}} & \multicolumn{1}{c}{\tiny{MSE}} & \multicolumn{1}{c}{\tiny{MSE}} & \multicolumn{1}{c}{\tiny{MSE}} & \multicolumn{1}{c}{\tiny{MSE}} \\
        \addlinespace[0.25em]
        Scale &
        \tiny{$10^{-8}$} & \tiny{$10^{-7}$} & \tiny{$10^{-8}$} & \tiny{$10^{-7}$} & \tiny{$10^{-8}$} & \tiny{$10^{-6}$} & \tiny{$10^{-6}$} & \tiny{$10^{-4}$} & \tiny{$10^{-3}$} & \tiny{$10^{-7}$} & \tiny{$10^{-4}$} & \tiny{$10^{-4}$} & \tiny{$10^{-4}$} & \tiny{$10^{-3}$} & \tiny{$10^{-3}$} & \tiny{$10^{-5}$} & \tiny{$10^{-4}$} & \tiny{$10^{-3}$} & \tiny{$10^{-6}$} & \tiny{$10^{-6}$} & \tiny{$10^{-1}$} & \tiny{$10^{-4}$} \\
        \midrule
        \method{Integral} &
        $\lightning$ & 10.9 & $\lightning$ & 95.2 & $\lightning$ & $\lightning$ & 239.9 & 28.0 & 25.9 & 117.6 & 2.1 & 12.7 & $\lightning$ & $\lightning$ & $\lightning$ & $\lightning$ & $\lightning$ & $\lightning$ & $\lightning$ & $\lightning$ & $\lightning$ & $\lightning$ \\
        \method{AD-Na\"{i}ve} &
        6.0 & $\lightning$ & 4.0 & $\lightning$ & 3.0 & $\lightning$ & 57.5 & 5.4 & 2.0 & 224.9 & 2.6 & 7.8 & 1.1 & 1.8 & 5.9 & 4.6 & 8.4 & 5.1 & 2.2 & 4.6 & 1.0 & 3.1 \\
        \method{AD-Reduc} &
        6.0 & $\lightning$ & 294.0 & 9.7 & $\lightning$ & $\lightning$ & 57.5 & 5.4 & 2.0 & 224.9 & 2.6 & 7.8 & $\lightning$ & $\lightning$ & $\lightning$ & 7.9 & 12.1 & 19.5 & 2.2 & 4.6 & 3.5 & 1.8 \\
        \method{Num-FD} &
        $\lightning$ & $\lightning$ & $\lightning$ & 51.3 & $\lightning$ & 8.5 & $\lightning$ & $\lightning$ & $\lightning$ & $\lightning$ & $\lightning$ & $\lightning$ & 2.7 & 10.3 & 47.0 & 7.1 & 18.9 & 8.5 & 53.8 & 10.9 & $\lightning$ & $\lightning$ \\
        \method{Num-FD-\compop} &
        174.0 & 1.9 & 19.0 & 9.3 & 192.0 & 1.8 & 6.2 & 3.4 & 2.6 & 5.2 & 1.6 & 3.9 & 1.2 & 8.4 & 22.2 & 4.6 & 20.1 & 7.0 & 3.4 & 4.6 & $\lightning$ & $\lightning$ \\
        \method{Num-Sm} &
        437.0 & 4.9 & $\lightning$ & $\lightning$ & $\lightning$ & $\lightning$ & 33.6 & 13.5 & 16.6 & 26.2 & 2.1 & 25.0 & $\lightning$ & $\lightning$ & $\lightning$ & $\lightning$ & $\lightning$ & $\lightning$ & $\lightning$ & $\lightning$ & $\lightning$ & $\lightning$ \\
        \method{Num-Sm-\compop} &
        267.0 & 21.0 & $\lightning$ & $\lightning$ & $\lightning$ & $\lightning$ & 30.6 & 6.6 & 8.0 & 79.0 & 1.8 & 7.1 & $\lightning$ & $\lightning$ & $\lightning$ & $\lightning$ & $\lightning$ & $\lightning$ & $\lightning$ & $\lightning$ & $\lightning$ & $\lightning$ \\
        \bottomrule 
    \end{tabular}
\end{table*}

\begin{table*}[]
    \centering
    \small
    \caption{Quantitative results for the use of learned antiderivatives of real-world signals in the task of \textbf{rendering}. We report results for two glossy BRDF lobes -- narrow and wide -- corresponding to Phong exponents $\alpha = 64$ and $\alpha = 8$, respectively. The symbol $\lightning$ indicates failure (see main text).}
    \label{tab:results-rendering}
    \begin{tabular}{lrrrrrrrrrrrr}
        \toprule
        \intcount &
        \multicolumn{6}{c}{1} & \multicolumn{6}{c}{2} \\
        \cmidrule(lr){2-7}
        \cmidrule(lr){8-13} 
        Spec. Lobe & \multicolumn{3}{c}{Narrow} & \multicolumn{3}{c}{Wide} & \multicolumn{3}{c}{Narrow} & \multicolumn{3}{c}{Wide} \\
        \cmidrule(lr){2-4}
        \cmidrule(lr){5-7}
        \cmidrule(lr){8-10}
        \cmidrule(lr){11-13}
        Metric & \multicolumn{1}{c}{\tiny{MSE}} & \multicolumn{1}{c}{\tiny{DSSIM}} & \multicolumn{1}{c}{\tiny{LPIPS}} & \multicolumn{1}{c}{\tiny{MSE}} & \multicolumn{1}{c}{\tiny{DSSIM}} & \multicolumn{1}{c}{\tiny{LPIPS}} & \multicolumn{1}{c}{\tiny{MSE}} & \multicolumn{1}{c}{\tiny{DSSIM}} & \multicolumn{1}{c}{\tiny{LPIPS}} & \multicolumn{1}{c}{\tiny{MSE}} & \multicolumn{1}{c}{\tiny{DSSIM}} & \multicolumn{1}{c}{\tiny{LPIPS}} \\
        \addlinespace[0.25em]
        Scale &
        \tiny{$10^{-1}$} & \tiny{$10^{-3}$} & \tiny{$10^{-3}$} & \tiny{$10^{-2}$} & \tiny{$10^{-3}$} & \tiny{$10^{-3}$} & \tiny{$10^{-2}$} & \tiny{$10^{-3}$} & \tiny{$10^{-3}$} & \tiny{$10^{-2}$} & \tiny{$10^{-3}$} & \tiny{$10^{-3}$} \\
        \midrule
        \method{Integral} &
        22.3 & 28.6 & 47.4 & 70.0 & 23.6 & 44.9 & $\lightning$ & $\lightning$ & $\lightning$ & $\lightning$ & $\lightning$ & $\lightning$ \\
        \method{AD-Na\"{i}ve} &
        1.6 & 6.8 & 8.4 & 2.7 & 4.5 & 7.6 & 1.4 & 5.1 & 7.8 & 1.7 & 3.3 & 5.9 \\
        \method{AD-Reduc} &
        1.6 & 6.8 & 8.4 & 2.7 & 4.5 & 7.6 & $\lightning$ & $\lightning$ & $\lightning$ & 4.4 & 12.7 & 23.2 \\
        \method{Num-FD} &
        3.6 & 15.8 & 31.1 & 4.3 & 6.7 & 11.3 & 7.8 & 16.5 & 33.2 & 1.9 & 7.7 & 14.5 \\
        \method{Num-FD-\compop} &
        21.7 & 26.1 & 52.1 & 72.0 & 22.8 & 46.0 & $\lightning$ & $\lightning$ & $\lightning$ & 133.2 & 19.7 & 47.1 \\
        \method{Num-Sm} &
        25.7 & 27.9 & 53.4 & 104.6 & 24.8 & 49.6 & $\lightning$ & $\lightning$ & $\lightning$ & $\lightning$ & $\lightning$ & $\lightning$ \\
        \method{Num-Sm-\compop} &
        20.7 & 25.1 & 49.9 & 76.9 & 22.7 & 45.9 & $\lightning$ & $\lightning$ & $\lightning$ & $\lightning$ & $\lightning$ & $\lightning$ \\
        \bottomrule 
    \end{tabular}
\end{table*}

\begin{table}[]
    \centering
    \small
    \renewcommand{\tabcolsep}{0.15cm}
    \caption{\textbf{Training time} in minutes on a single A40 GPU for each combination of signal modality, integration order, and method. The symbol $\lightning$ indicates failure of training to converge.}
    \label{tab:training-time}
    \begin{tabular}{lrrrrrrr}
        \toprule
        Modal. \tiny{($\indimcount \rightarrow \outdimcount$)} & \multicolumn{3}{c}{Motion \tiny{($1 \rightarrow 69$)}} & \multicolumn{2}{c}{Image \tiny{($2 \rightarrow 3$)}} & \multicolumn{2}{c}{SDF \tiny{($3 \rightarrow 1$)}} \\
        \cmidrule(lr){2-4}
        \cmidrule(lr){5-6}
        \cmidrule(lr){7-8}
        \intcount & \multicolumn{1}{c}{1} & \multicolumn{1}{c}{2} & \multicolumn{1} {c}{3} & \multicolumn{1}{c}{1} & \multicolumn{1}{c}{2} & \multicolumn{1}{c}{1} & \multicolumn{1}{c}{2} \\
        \midrule
        \method{Integral} &
        3.8 & 3.4 & 2.5 & 6.4 & 4.7 & 36.3 & 32.8 \\
        \method{AD-Na\"{i}ve} &
        4.0 & 53.8 & 108.7 & 22.3 & 478.6 & 48.1 & 836.2 \\
        \method{AD-Reduc} &
        4.0 & 3.7 & 5.5 & 22.3 & 81.7 & 48.1 & 33.8 \\
        \method{Num-FD} &
        1.7 & 3.5 & 5.9 & 10.8 & 13.4 & 13.7 & 19.1 \\
        \method{Num-FD-\compop} &
        5.0 & 4.6 & 11.8 & 19.8 & 30.7 & 99.7 & 103.0 \\
        \method{Num-Sm} &
        61.9 & 24.4 & $\lightning$ & 107.2 & $\lightning$ & 209.0 & $\lightning$ \\
        \method{Num-Sm-\compop} &
        76.0 & 22.3 & $\lightning$ & 110.0 & $\lightning$ & 347.9 & $\lightning$ \\
        \bottomrule 
    \end{tabular}
\end{table}

\section{Summary, Recommendations, and Conclusion}
\label{sec:conclusion}

We have introduced a hierarchical taxonomy of supervision structures for learning repeated antiderivatives from function samples.
In particular, we contrasted direct supervision based on antiderivative estimates with several method classes that rely on repeated differentiation, including both automatic and numerical approaches.
As part of this framework, we explored integral reduction techniques as a means to mitigate the computational overhead associated with repeated integration or differentiation.

Our systematic experimental analysis of all presented methods across a diverse set of modalities reveals that differential supervision via na\"{i}ve automatic differentiation generally outperforms all competing approaches in terms of result quality.
However, this performance comes at the cost of substantial training time, particularly for higher-order integration.
A practical alternative in many scenarios is numerical differentiation via finite differences, which often achieves a reasonable trade-off between quality and computational cost.
Nevertheless, it is important to incorporate a compensation operator to correct for the signal smoothing inherent in this differentiation scheme.
Our findings suggest directions for future research, such as investigating progressive supervision paradigms that leverage different supervisory signals throughout the learning process.

Interestingly, our analysis reveals that antiderivative quality correlates only weakly with downstream task performance.
Cumulative schemes often excel at efficiently computing highly non-local aggregates, where local inaccuracies tend to cancel out.
Certain combinations of signal modality and integration order therefore benefit from more sophisticated integral reduction schemes.

We believe our work provides a foundation for integrating classical cumulative operators into modern neural pipelines, and we hope the insights gained will carry over to a broader set of learning tasks involving differential and integral operators.


\printbibliography                

\end{document}